\documentclass[10pt,twocolumn,letterpaper]{article}

\usepackage{3dv}
\usepackage{times}
\usepackage{epsfig}
\usepackage{graphicx}
\usepackage{amsmath}
\usepackage{amssymb}

\usepackage{wrapfig}

\usepackage[pagebackref=true,breaklinks=true,letterpaper=true,colorlinks,bookmarks=false]{hyperref}

\threedvfinalcopy 

\def\threedvPaperID{****} 

\ifthreedvfinal\fi
\begin{document}

\title{Synthesizing Training Images for Boosting Human 3D Pose Estimation}

\author{Wenzheng Chen$^{1}$
\and
Huan Wang$^{1}$
\and
Yangyan Li$^{2}$
\and
Hao Su$^{3}$
\and
Zhenhua Wang$^{4}$
\and
Changhe Tu$^{1}$
\and
Dani Lischinski$^{4}$
\and
Daniel Cohen-Or$^{2}$
\and
Baoquan Chen$^{1}$\\\\
\and
$^{1}$ Shandong University
$^{2}$ Tel Aviv University
$^{3}$ Stanford University
$^{4}$ Hebrew University
}

\maketitle

\begin{abstract}
Human 3D pose estimation from a single image is a challenging task with numerous applications. Convolutional Neural Networks (CNNs) have recently achieved superior performance on the task of 2D pose estimation from a single image, by training on images with 2D annotations collected by crowd sourcing. This suggests that similar success could be achieved for direct estimation of 3D poses. However, 3D poses are much harder to annotate, and the lack of suitable annotated training images hinders attempts towards end-to-end solutions.
To address this issue, we opt to automatically synthesize training images with ground truth pose annotations. Our work is a systematic study along this road. We find that pose space coverage and texture diversity are the key ingredients for the effectiveness of synthetic training data. We present a fully automatic, scalable approach that samples the human pose space for guiding the synthesis procedure and extracts clothing textures from real images. Furthermore, we explore domain adaptation for bridging the gap between our synthetic training images and real testing photos. We demonstrate that CNNs trained with our synthetic images out-perform those trained with real photos on 3D pose estimation tasks.

\end{abstract}
\vspace{-0.5cm}

\section{Introduction}
\label{sec:intro}


Recovering the 3D geometry of objects in an image is one of the longstanding and most fundamental tasks in computer vision. In this paper, we address a particularly important and challenging instance of this task: the estimation of human 3D pose from a single (monocular) still RGB image of a human subject, which has a multitude of applications~\cite{jain2010movie,zhou2010parametric,weiss2011,guan2012drape}.

Most of the existing work in human pose estimation produces a set of 2D locations corresponding to the joints of an articulated human skeleton~\cite{mori2002estimating}.
An additional processing stage is then required in order to estimate the 3D pose from 2D joints~\cite{taylor2000reconstruction,anguelov2005scape,guan2009estimating,ramakrishna2012reconstructing,fan2014pose,akhter2015pose,reviewer1_1,yash2016,WangWLYG14}.
However, errors are accumulated in this two-stage 3D pose estimation system. Inspired by the recent success of training CNNs in an end-to-end fashion,  one might expect that direct estimation of 3D poses should be more effective.

In this paper, we directly estimate 3D poses, with a focus on synthesis of effective training data for boosting the performance of deep CNN networks. This task of direct 3D pose estimation is more challenging than the 2D case since one needs a large number of human bodies with different genders and fitness levels, seen from a wide variety of viewing angles, featuring a diversity of poses, clothing, and backgrounds. Therefore, an effective CNN has to be trained by a large number of training examples, which cover well the huge space of appearance variations. \emph{Obtaining a sufficiently diverse training set of 3D groundtruth is a major bottleneck.} Crowd-sourcing is not a practical option here, because manually annotating a multitude of images with 3D skeletons by human workers is not a feasible task: the annotations must be marked in 3D, and, furthermore, it is inherently hard for humans to estimate the depth of each joint given only a single 2D image. Massive amounts of 3D poses may be captured by Motion Capture (MoCap) systems, however they are not designed to capture the accompanying appearance, thus it is difficult to achieve the necessary diversity of the training data.

\begin{figure}[t]
	\vspace{-0.5cm}
	\begin{center}
		\includegraphics[width=1.0\linewidth]{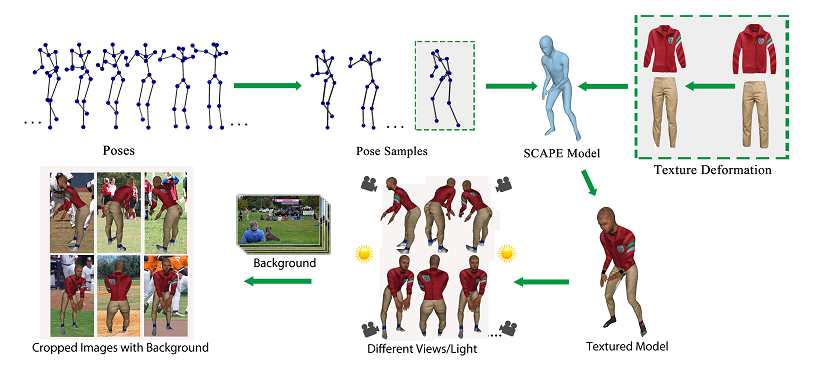}
	\end{center}
	\vspace{-0.5cm}
	\caption{Our training data generation pipeline. The 3D pose space is sampled and the samples are used for deforming SCAPE models. Meanwhile, various clothes textures are mapped onto the human models. The deformed textured models are rendered using a variety of viewpoints and light sources, and finally composited over real image backgrounds.
	}
	\vspace{-0.5cm}
	\label{fig:pipeline}
\end{figure}

Recently, synthesized images have been shown to be effective for training CNNs~\cite{su2015render,massa2015deep}.  In this work, we also use synthetic images to address the bottleneck of training data.  However, the challenges we face are rather different and much more difficult than existing works, which mostly target on man-made objects, where rigid transformations are applied to generate pose variation, and very limited work has been done to address texture variation. Unlike static man-made objects, human bodies are non-rigid, richly articulated, and wear varied clothing.  

We systematically explore these issues in this paper, with an emphasis on driving the synthesis procedure by real data and bridging the domain gap between synthetic and real data. We build a statistical model from a large number of 3D poses from a MoCap system, or inferred from human annotated 2D poses, from which we can sample as many body types and poses as needed for training. We further present an automatic method for transferring clothing textures from real product images onto a human body. Without complicated physical simulation as in traditional cloth modeling~\cite{Baraff98Large}, this data-driven texture synthesis approach is highly scalable, but still retains visual details such as wrinkles and micro-structures. Effectively, we generate 10,556 human models with unique and quality textured clothing. Given a sampled (articulated) pose, we render the textured human body, overlaid on a randomly chosen background image to generate a synthetic image. In our experiments we generated 5,099,405 training images.

Admittedly, there is a visual gap between synthetic training data and real testing data, as a result, the features between them live in different domains. To make best of our synthetic data, we employ \emph{domain adaptation}~\cite{bendavid2010atheory} to bridge the gap by shifting two domains towards one common space. We design a new strategy for domain adaptation, and show that our new strategy can better bridge two domains.

Our synthetic training data and the code to generate it will be made publicly available\footnote{The code, data and mdoel can be found at \url{http://irc.cs.sdu.edu.cn/Deep3DPose/}}. We anticipate it to be impactful for the community, due to its unprecedented scale and diversity, compared to any of the existing datasets with ground truth 3D annotations, which results in better performance in 3D pose estimation. To demonstrate this, we train several state-of-the-art CNNs with our synthetic images, and evaluate their performance on the human 3D pose estimation task using different datasets, observing consistent and significant improvements over the published state-of-the-art results. In addition to the large-scale synthetic dataset, we also created Human3D+, a new richer dataset of images with 3D annotation. This dataset will be made public, as well.

\section{Related Work}
\label{sec:related}


Analyzing human bodies in images and videos has been a research topic for many decades, with particular attention paid to estimation of human body poses~\cite{hogg1983model,lee1985determination}. While some earlier works were based on local descriptors \cite{mori2006recovering,ferrari08progressive}, the recent emergence of CNNs has led to significant improvements in body pose estimation from a single image.

\paragraph{Human Pose Datasets.} FLIC~\cite{sapp2013modec}, MPI~\cite{andriluka14cvpr}, and LSP~\cite{johnson10clustered,johnson11learning} are the largest available datasets, which have 5003, 2179 and 2000 fully annotated human bodies respectively. Generally speaking, CNNs can extract high quality image features, and these features can be adapted for various other vision tasks. However, fine-tuning a CNN for a specific task, such as pose estimation, still requires a large number of annotated images. Existing datasets listed above are still too limited in scale and diversity.

\paragraph{2D Pose Estimation.} Toshev and Szegedy~\cite{toshev2014deeppose} proposed a cascade of CNN-based regressors for predicting 2D joints in a coarse to fine manner. Both Fan et al.~\cite{fan2015combining} and Li et al.~\cite{li2014heterogeneous} proposed to combine body-part detection and 2D joints localization tasks.
Gkioxari et al.~\cite{gkioxari2014rcnns} also explored a multi-task CNN, where an action classifier is combined with a 2D joints detector. Both Tompson et al.~\cite{tompson2014joint} and Chen and Yuille~\cite{chen2014articulated} proposed to represent the spatial relationships between joints by graphical models, while training a CNN for predicting 2D joint positions.
Pishchulin et al.~\cite{941} jointly solve the tasks of detection and pose estimation in a multi-people image. Wei et al.~\cite{wei2016cpm} add CNN to pose machine framework to predict 2D joints.

\paragraph{3D Pose Estimation.} Since all of the above methods estimate 2D poses, several methods have been proposed for recovering 3D joints from their 2D locations
\cite{taylor2000reconstruction,ramakrishna2012reconstructing,fan2014pose,akhter2015pose,reviewer1_1,yash2016,WangWLYG14}. However, such methods operate only the 2D joint locations, ignoring all other information in the input image, which might contain important cues for inferring the 3D pose. CNN solutions are likely to work better, since they are capable of taking advantage of additional information in the image. We found that CNNs trained with our synthetic images outperform 2D-pose-to-3D-pose methods, even when the latter are provided with the ground truth 2D joint locations, not to mention when they start from automatically estimated 2D poses.

Li and Chan~\cite{li20143d,li2015maximum} proposed CNNs for 3D pose estimation trained using the Human3.6M dataset~\cite{ionescu2011latent,h36m_pami}, where the ground truth 3D poses were captured by a MoCap system. Their method achieves high performance on subjects from the same dataset that were put aside as test data. However, we found that the performance of their CNN drops significantly when tested on other datasets, which indicates a strong overfit on the Human3.6M dataset. The reason for this may be that while there are millions of frames and 3D poses in this dataset, their variety is rather limited.

\paragraph{Human Pose Data Synthesis.} Several recent works synthesize human body images from 3D models for training algorithms. However, these works are limited in scalability, pose variation or viewpoint variation.  Both Pishchulin et al.~\cite{pin2012syn} and Zhou et al.~\cite{zhou2010parametric} fit 3D models to images, and deform the model to synthesize new images. However, they either request user to supply a good 3D skeleton and segmentation, or need considerable user interaction. Vazquez et al.~\cite{vaz2014game} collect synthesized images with annotations from game engines, thus it is restricted to certain scenes and people. Park et al.~\cite{park2015video} use layering to reconstruct images with different pose, leading to imprecise and poor resolution synthetic images. Since none of the above methods can generate large-scale training images, they can hardly satisfy the demand of CNNs.

Other methods exist that recover 3D pose by adding extra information beside 2D annotations. Agarwal et al.~\cite{Agarwal20063dpose} predict 3D pose from silhouette. Radwan et al.~\cite{Radwan20133dpose} use both kinematic and orientation constraints to estimate self-occlusion 3D pose. Zhou et al.~\cite{xiaowei2016} and Tekin et al.~\cite{Tekin_2016_CVPR} use monocular video to recover 3D pose. Ionescu et al.~\cite{IonescuCS14} recover 3D pose with depth information. However, the additional constraints may also introduce error and decrease the reliability of the full system.

\paragraph{Domain Adaptation.} Most previous methods in domain adaptation worked on a fixed feature representation~\cite{LiDXT14}. Recently, there is a trend to combine domain adaptation and deep feature learning in one training process. Ganin and Lemptisky~\cite{ganin2015unsupervised} proposed a new deep network architecture to perform domain adaptation by standard backpropagation training.
Independently, Ajakan \etal~\cite{Ajakan14} proposed a similar deep network to learn feature representation that is predictive for the classification task in source domain but uninformative to distinguish source and target domain. Rather than confusing a domain classifier, Tzeng \etal~\cite{Tzeng14} introduced an adaptation layer along with a domain confusion loss to minimize the distance between feature distributions of two domains by maximum mean discrepancy. Long \etal~\cite{Long15} extended~\cite{Tzeng14} by using multiple kernel discrepancy and applying it on all the task specific layers.

\section{Training a 3D Human Pose Estimator by Synthetic Data}
\label{sec:synthesis}

Training effective CNNs for 3D pose estimation requires a large number of 3D-annotated images. Our synthesis based approach can generate an infinite number of possible combinations of viewing angles, human poses, clothing articles, and backgrounds by  combining these properties. The generated combinations should be chosen such that (i) the result resembles a real image of a human (we refer to this as the \textbf{alignment principle}; and (ii) the synthesized images should be diverse enough to sample well the space of real images of humans (the \textbf{variation principle}).

\begin{figure}[t]
    \vspace{0.2cm}
	\begin{center}
		\includegraphics[width=0.8\linewidth]{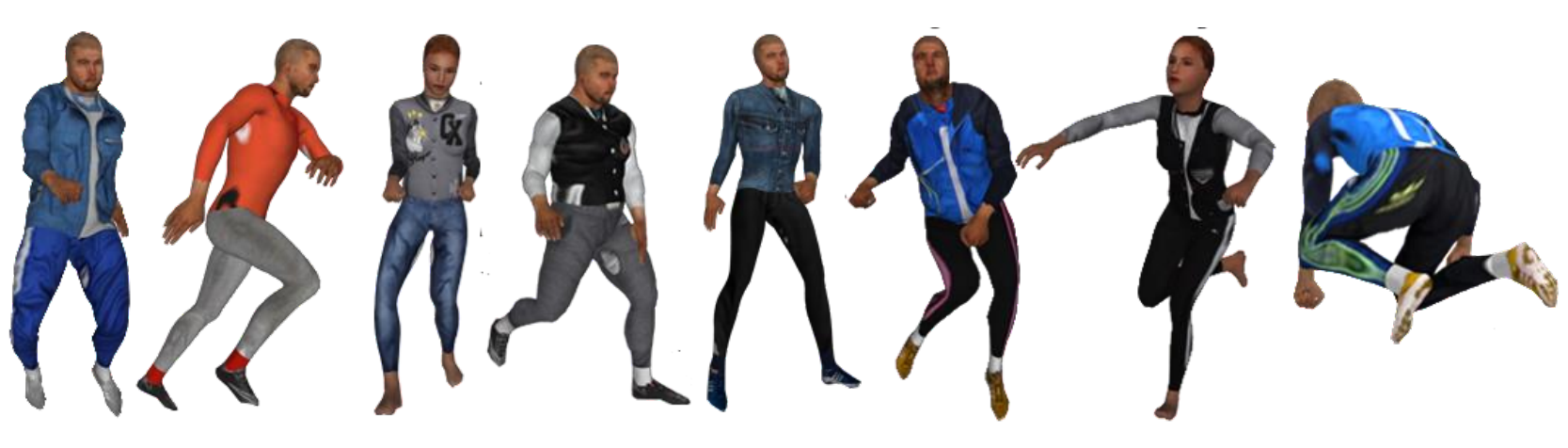}
	\end{center}
	\vspace{-0.4cm}
	\caption{A sample of synthesized 3D human models.
	}
	\vspace{-0.5cm}
	\label{fig:3d_human_model}
\end{figure}

Our training data generation approach is illustrated in Figure~\ref{fig:pipeline}, which consists of sampling the pose space (Section~\ref{sec:pose}), and using the results to generate a large collection of articulated 3D human models of different body types with SCAPE~\cite{anguelov2005scape}. The models are textured with realistic clothing textures extracted from real images (Section~\ref{sec:texture}). A sample of synthesized 3D human models in various body types, clothes and poses is shown in Figure~\ref{fig:3d_human_model}. Then, the textured models are rendered and composited over real image backgrounds (Section~\ref{sec:rendering_composition}).

Besides improving realism of the synthetic images in the image generation pipeline, domain adaptation method can be applied to bridge the two domains in the feature space. Inspired by~\cite{ganin2015unsupervised}, we train a CNN model such that the features extracted by this model from synthetic images and real images share a common domain.


\subsection{Body Pose Space Modelling}
\label{sec:pose}

\begin{figure}[t]
	\vspace{-0.5cm}
	\begin{center}
		\includegraphics[width=0.8\linewidth]{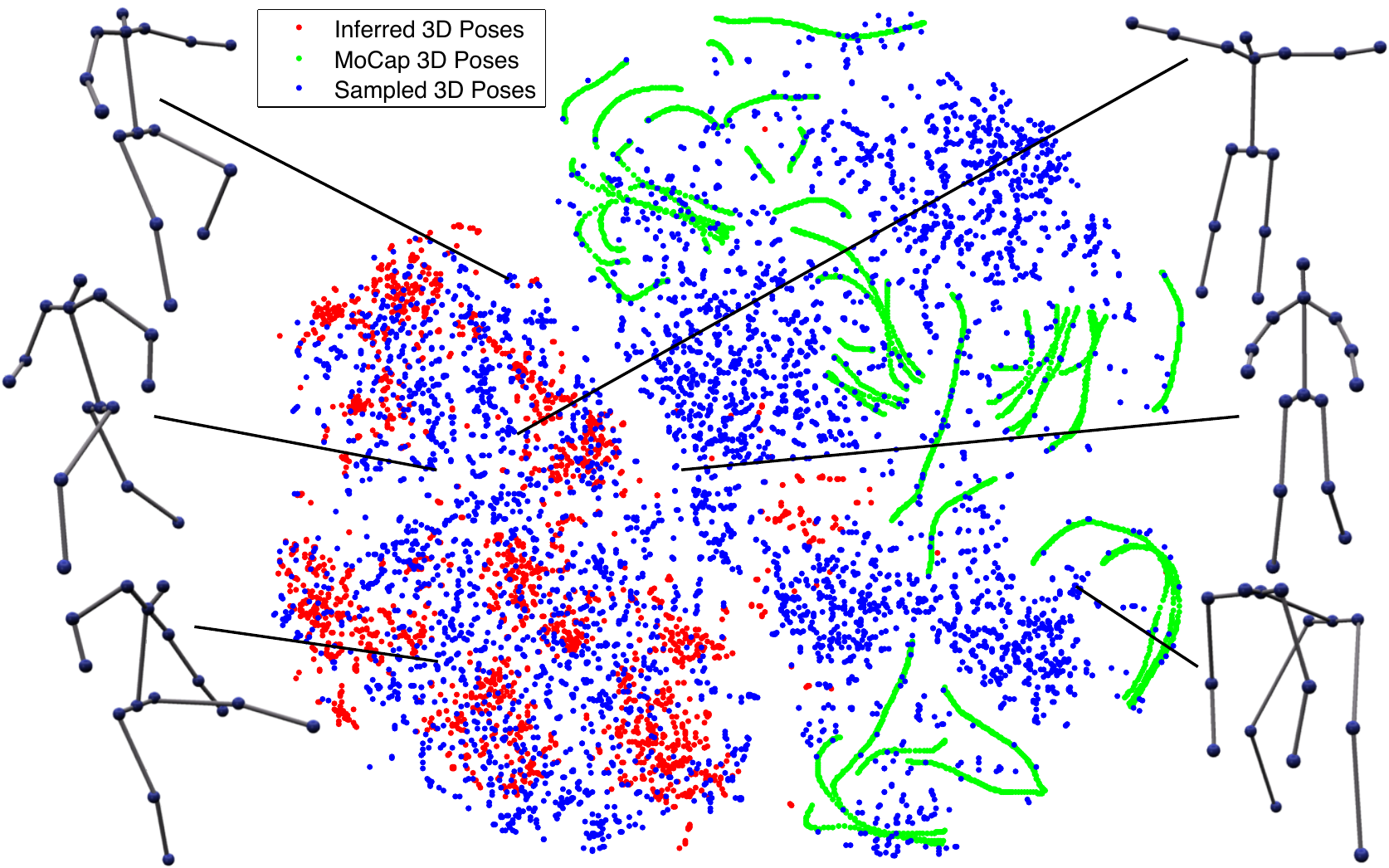}
	\end{center}
	\vspace{-0.5cm}
	\caption{A sample of poses drawn from the learned non-parametric Bayesian network and t-SNE 2D visualization of the high dimensional pose space. Note that the 3D poses inferred from human annotated 2D poses (red) are complementary to MoCap 3D poses (green). New poses (blue) can be sampled from the prior learned from both the MoCap and inferred 3D poses, and have better coverage of the pose space.}
	\vspace{-0.5cm}
	\label{fig:pose_space}
\end{figure}

The pose distribution of the synthetic images should agree with that of real-world images. 
In case there are enough poses available that cover the entire pose space, we can keep sampling poses from this pool, and select a large number of poses whose distribution approximates that of real images well. Unfortunately, we found the poses from the existing datasets only sparsely cover a small portion of the pose space. The existing datasets, e.g. CMU MoCap dataset~\cite{mocap_cmu} and Human3.6M~\cite{h36m_pami}, are classified by different actions. Even though they cover many common human actions, they can hardly represent the entire pose space.

To better cover the pose space, unseen poses should also be generated. The key challenge is to make sure the generated unseen poses are valid. The idea is to learn the variations of parts that frequently occur together and produce new poses by combining these parts. We learn a sparse and non-parametric Bayesian network from a set of input poses, to factorize the pose representation, and then composite substructures for generating new poses, as proposed in~\cite{lehrmann2013nbn}. Pose samples drawn from the learned Bayesian network exhibit richer variations due to the substructure composition; meanwhile, the poses stay valid as substructures are composited only when appropriate.

\begin{figure}[t]
	\vspace{-0.5cm}
	\begin{center}
		\includegraphics[width=0.8\linewidth]{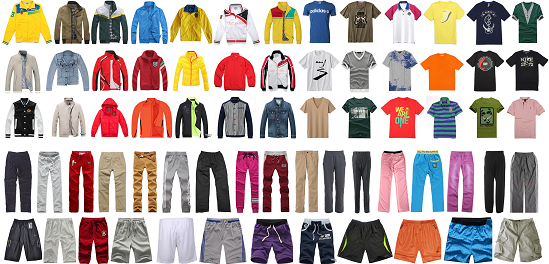}
	\end{center}
	\vspace{-0.4cm}
	\caption{A sample of clothing images used for transferring texture onto 3D human models. They are from Google Image Search.}
	\vspace{-0.5cm}
	\label{fig:clothes_gallery}
\end{figure}

We learn the Bayesian network from both MoCap 3D poses and 3D poses inferred from human annotated 2D poses. We use the CMU MoCap dataset because it contains more types of actions than Human3.6M.
We use LSP~\cite{johnson10clustered,johnson11learning} as a 2D pose source, and Akhter et al.~\cite{akhter2015pose} to recover 3D poses from 2D annotations. The complementary nature of MoCap 3D poses and inferred 3D poses is demonstrated in Figure~\ref{fig:pose_space}.

Note that the poses sampled from the learned Bayesian network cover the input MoCap and inferred 3D poses well. Moreover, since the prior is learned from both the MoCap and inferred 3D poses, the ``interpolation'' between the MoCap and inferred poses can be sampled from the learned Bayesian network as well, due to the compositionality.

Each sample of the pose space yields a set of 3D joints. The 3D joints, together with other parameters, such as gender and fitness level, are provided as input to SCAPE~\cite{anguelov2005scape} for yielding richly varied articulated human models. The fitness levels are supplied based on an empirical distribution, though that learnt from real data might be even better.

\subsection{Clothing Texture Transfer}
\label{sec:texture}

Humans wear a wide variety of clothing. In the real world, clothes are
in a wide diversity of appearances. Our goal is to synthesize clothed humans whose appearance mimics that seen in real images. However, it is hard to design or learn a parametric model for generating suitable textures. Instead, we propose a fully automatic light-weighted  approach that transfers large amount of clothes textures from images onto human 3D models.

\begin{figure}[t]
	\vspace{-0.5cm}
	\begin{center}
		\includegraphics[width=0.8\linewidth]{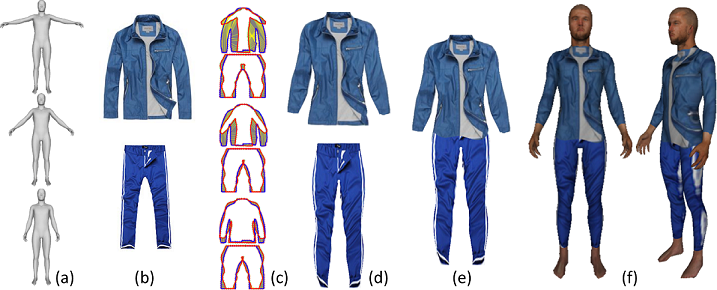}
	\end{center}
	\vspace{-0.4cm}
	\caption{Contour matching for clothing texture transfer. We render 3D human models in a few candidate poses (a), and try to match their contour to those in real clothing images (b). Next, textures from real images are warped according to the best contour matching (c,d,e) and projected onto the corresponding 3D model. Finally, the textures are mirrored (left-right as well as front-back) to cover the entire 3D model (f). Note that the winkles on clothes of the 3D body are transferred from product images and are still relatively natural.
	}
	\vspace{-0.5cm}
	\label{fig:texture_transfer}
\end{figure}

There are many product images of clothes available online, in which clothes are often imaged in canonical poses with little or no foreground occlusion or background clutter (see Figure~\ref{fig:clothes_gallery}), thus can be easily transferred from an image onto a 3D model. Our approach is to collect and analyze such images, and then use them to transfer realistic clothing textures onto our 3D human models. The transfer is done by establishing a matching between the contour of clothing article in the image and the corresponding part of a rendered human model (see Figure~\ref{fig:texture_transfer}). In general, we can transform clothes which fit human body, like tracksuits, skirts, coat, etc.

Firstly, we collect a large set of images of sportswear of various styles, whose backgrounds are simple and can be automatically segmented out.
This results in 2,000 segmented images (1,000 for upper body and 1,000 for lower body). Correspondingly, we split a 3D human model into overlapping upper and lower parts for matching. These two parts are projected onto the clothing images.  We use continuous dynamic time warping~\cite{munich1999continuous} for computing the dense correspondences $M(p)$ between the contours of the two human body parts $\mathcal{P}=\{p_i\}$ and those of the imaged clothing articles.

Once the dense correspondences between the contours are available, we warp the image of the article to fit the projected contour, and the warped image is then used to define the texture for the corresponding portion of the 3D human model. MLS image deformation~\cite{schaefer2006image} is applied for a smooth warping. The resulting textures on the 3D model are mirrored both left-right and front-back for better model coverage. The seam between upper and lower body is perturbed for generating more variations.


We found that the clothing images can be matched better, and result in less deformation when the 3D human model is provided in multiple candidate poses, and we pick the pose that results in minimal deformation.
In practice, we picked three upper body candidate poses. Note that rather than pursuing realistic clothing effect, we try to generate human 3D models of high diversity to prevent CNNs from picking up unreliable patterns.

The head, feet, and hands, which may not be covered by clothes, are texture mapped with a set of head, shoes and skin textures. Their colors are further perturbed by blending to generate more variations before clothing textures are transferred onto the models. Since the area of these regions is relatively small, their appearance is less important than the clothes, thus we opt for this simple strategy that is also scalable.

\subsection{Rendering and Composition}
\label{sec:rendering_composition}

\begin{figure}[t]
	\vspace{0.2cm}
	\begin{center}
		\includegraphics[width=0.75\linewidth]{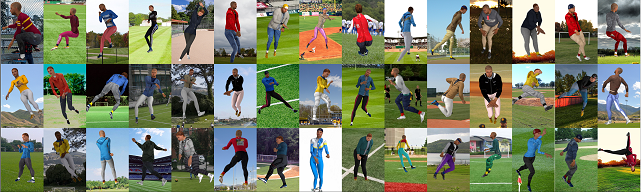}
		\includegraphics[width=0.75\linewidth]{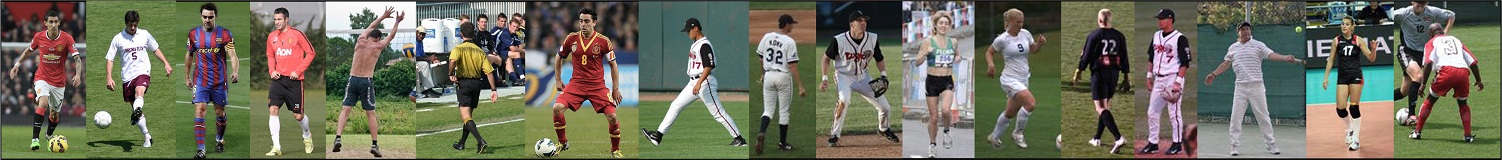}
	\end{center}
	\vspace{-0.4cm}
	\caption{A sample of synthetic training images (3 top rows) and real testing images (bottom row). The synthetic images may look fake to a human, but exhibit a rich diversity of poses and appearance for pushing CNNs to learn better.}
	\vspace{-0.5cm}
	\label{fig:synthetic_vs_real}
\end{figure}

Finally, textured human models in various poses are ready to be rendered and composited into synthetic images for CNN training. Three factors are important in the rendering process: camera viewpoint, lighting, and materials. The camera viewpoint is specified with three parameters: elevation, azimuth, and in-plane rotation. Typically, perturbations are added to the in-plane rotation parameter, by rotating the training images, to augment and generate more training data. Perturbations can also be added to the elevation and azimuth parameters. Starting from the camera viewpoint associated with each 3D pose, we add Gaussian perturbations with standard deviations of $15$, $45$, and $15$ degrees to the elevation, azimuth and in-plane rotation parameters, respectively. Various lighting models, number and energy of light sources are used during the rendering. The color tone of the body skin is also perturbed to represent different types of skin colors. Each rendered image is composited over a randomly chosen sports background image. We collected $796$ natural images from image repositories and search engines to serve as background. As shown in Figure~\ref{fig:synthetic_vs_real}, the synthetic images exhibit a wide variety of clothing textures, as well as poses, and present comparable complexity to real images.

\begin{figure}[t]
	\vspace{0.2cm}
	\begin{center}
        \includegraphics[width=0.3\linewidth]{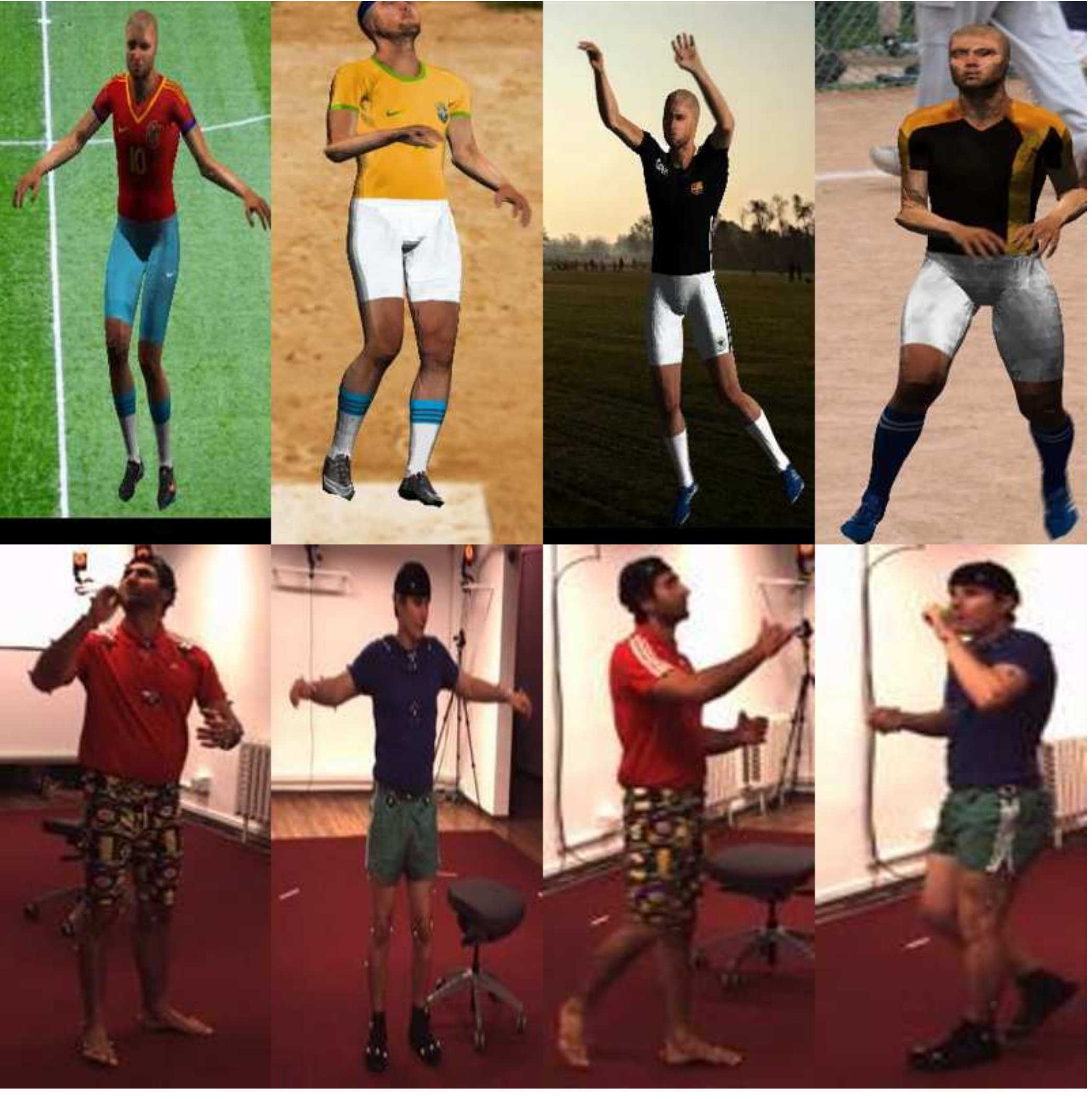}
		\includegraphics[width=0.4\linewidth]{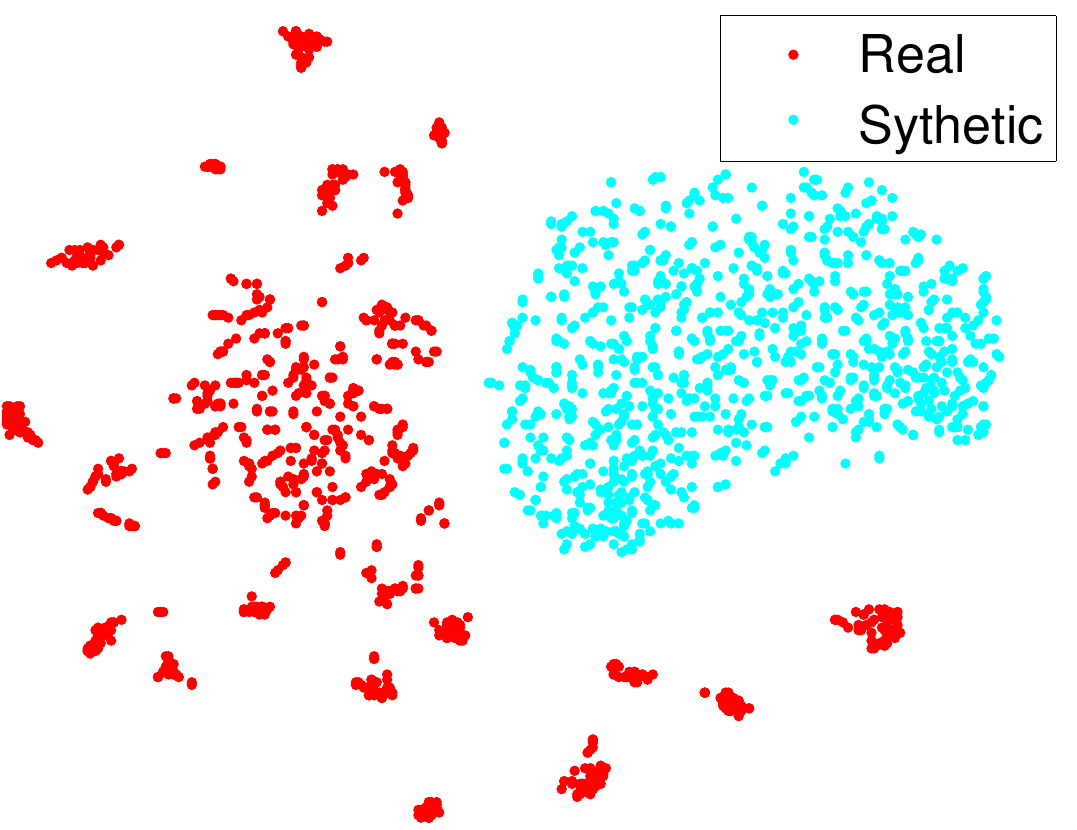}
	\end{center}
	\vspace{-0.4cm}
	\caption{Visual differences between synthetic images and real images form Human 3.6M (left), and t-SNE 2D visualization of AlexNet pool5 features (right). We can see a clear gap between the AlexNet pool5 features. Note that there are many clusters in the distribution of real images, this is because images in Human 3.6M are quite similar and every cluster is an actor doing similar actions. Moreover, the distribution of synthetic images is only in one cluster and more average, which shows that synthetic images are more diverse.}
	\vspace{-0.5cm}
	\label{fig:synthetic_vs_real2}
\end{figure}

\subsection{Domain Adaptation}
\label{sec:domain_adaptation}

Ideally, training images and testing images should live in the same domain, such that the model learned from training images can be directly applied on testing images. However, to seize the desired amount of training images with affordable computation and modeling, the realism of our synthetic images is compromised, inevitably. As a result, there is a gap between the CNN features of synthetic images and real images (see Figure~\ref{fig:synthetic_vs_real2}). Such gap hurts the performance on real images of the models trained from the synthetic images.

\begin{figure}[t]
	\vspace{-0.5cm}
	\begin{center}
		\includegraphics[width=0.8\linewidth]{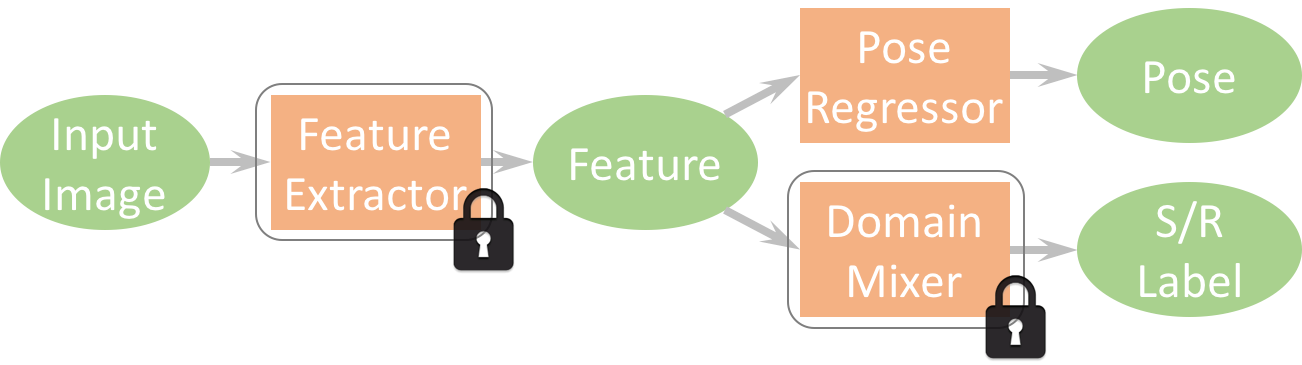}
	\end{center}
	\vspace{-0.5cm}
	\caption{Our domain adaptation network is composed of feature extractor, pose regressor and domain mixer. A two stage training scheme for learning effective models from synthetic and real images, where either the feature extractor or the domain mixer is fixed.}
	\vspace{-0.5cm}
	\label{fig:network}
\end{figure}

We train a domain adaptation network to map synthetic and real data to the same domain.
More specifically, we utilize a domain adaptation network that is composed of three components: feature extractor, pose regressor and domain mixer (see Figure~\ref{fig:network}). The feature extractor is responsible of transfering images into a feature space, where any typical CNN models can serve for this goal. In our case, we adopt AlexNet (until pool5 layer) as our feature extractor. The extracted features are sent to pose regressor for serving the 3D pose estimation task, as well as to domain mixer for bridging the feature gap between synthetic images and real images. The domain mixer aims at pushing features from synthetic images and real images into a common domain. We utilize an adversarial training for realizing this goal. The domain mixer components in the domain adaptation network actually is a synthetic vs. real classifier, which, given the feature of an input image, predicts whether the features are from a synthetic image or real image.

The input to our domain adaptation network are synthetic images, for which pose annotations are available, and real images, with or without pose annotations. We train our domain adaptation network in two stages.  More conceretley, we formulate the network loss as:

\begin{align}
Loss &= L_{reg} + L_{domain}\\
L_{reg} &= \sum \left \| p_{pre} - p_{gt} \right \|_{2}
\end{align}
where $Loss$ is the whole loss of domain adaptation network.  $L_{reg}$ is the regression loss, an Euclidean loss between the predicted pose and ground truth pose. $L_{domain}$ is the domain adaptation loss, which is different in 2 stages, as shown below. Here $p_{x}$ is the possibility of x being real image.

\begin{align}
L_{domain}^{stage1}=-\sum_{x\in real}logp_{x}-\sum_{x\in synthetic}log(1-p_{x})\\
L_{domain}^{stage2}=-\sum_{x}(0.5logp_{x}+0.5log(1-p_{x}))
\end{align}

We describe here how the two-stage training scheme works. In the first stage, we fix parameters in the feature extractor to train the pose regressor for 3D pose prediction and the domain mixer for real and synthetic images distinction. Here $d_{pre}$ is the predicted domain label and $d_{gt}$ is the ground truth label, with synthetic data label being $0$ and real data being $1$. After this stage, we  obtain a strong pose regressor and a synthetic-real classifier.
In the second stage, we fix the parameters of the domain mixer to train the feature extractor that would not only benefit the pose regressor but also confuse the synthetic-real classifier. More specifically, we confuse the synthetic-real classifier by enforcing it to output a probability distribution of (0.5, 0.5) for all synthetic and real images. Here, $d_{0.5,0.5}$ is the confused label where both synthetic data and real data are $0.5$. Such confusion indicates that the feature extractor has been adapted to generate indistinguishable features for these two domains.

Note that we train the network in a circular way, stage1 is alternate with stage2. This is because domain classifier is based on current feature extractor. If we modify the extractor in stage2, then we need to refine new classifier in stage1. So it is a circular process.


Our method has two differences from \cite{ganin2015unsupervised}. Firstly, we use the $50\%-50\%$ classification probability objective to yield a common domain, rather than the gradient reverse approach that encourages a common domain by the ``misleading'' gradients.  The advantage of this formulation is also reported in~\cite{tzeng2015simultaneous}.
Secondly, we train the network in an adversarial two-stages scheme. We show that the performances are better than~\cite{ganin2015unsupervised} with these two designs. Note that, the pose annotations of real images are not necessary for training our network. We show that the domain adaptation network is especially useful in the situation when real data is rare.

\section{Results and Discussion}
\label{sec:results}

We first introduce the datasets for evaluating 3D pose estimation in Section~\ref{sec:datasets}. Then we demonstrate the effectiveness of our synthetic training data in 3D pose estimation task by feeding it into a number of different CNNs in Section~\ref{sec:3d_pose}. Then we show how domain adaptation can be applied for better utilizing our synthetic data in Section~\ref{sec:domain_adaptation_results}. We study the performance of our synthetic datasets with some additional experiments in Section~\ref{sec:control_experiments}.

\subsection{Evaluation Datasets}
\label{sec:datasets}

The lack of images with 3D pose annotations is not only posing a problem for training, but also for evaluating 3D pose estimation methods. Existing datasets with 3D pose annotations, such as Human3.6M~\cite{h36m_pami} and HumanEva~\cite{mocap_humaneva}, have been captured in controlled indoor scenes, and are not as rich in their variability (clothing, lighting, background) as real-world images of humans.

Thus, we have created Human3D+, a new richer dataset of images with 3D annotation, captured in both indoor and outdoor scenes such as room, playground, and park, and containing general actions such as walking, running, playing football, and so on. The dataset consists of 1,574 images, captured with Perception Neuron MoCap system by Noitom Ltd.~\cite{mocap_noitom}. These images are richer in appearance and background, better representing human images in real-world scenarios, and thus are better suited for evaluating 3D pose estimation methods\footnote{The sensors mounting strips are artificial, but necessary for accurate capturing. However, since such strips do not appear in Human3.6M or in our synthetic images, it is not harmful for the comparison fairness.}. See our supplementary material for a sample of the images from our evaluation dataset.


\begin{figure}[t]
	\vspace{-0.5cm}
	\begin{center}
		\includegraphics[width=0.4\linewidth]{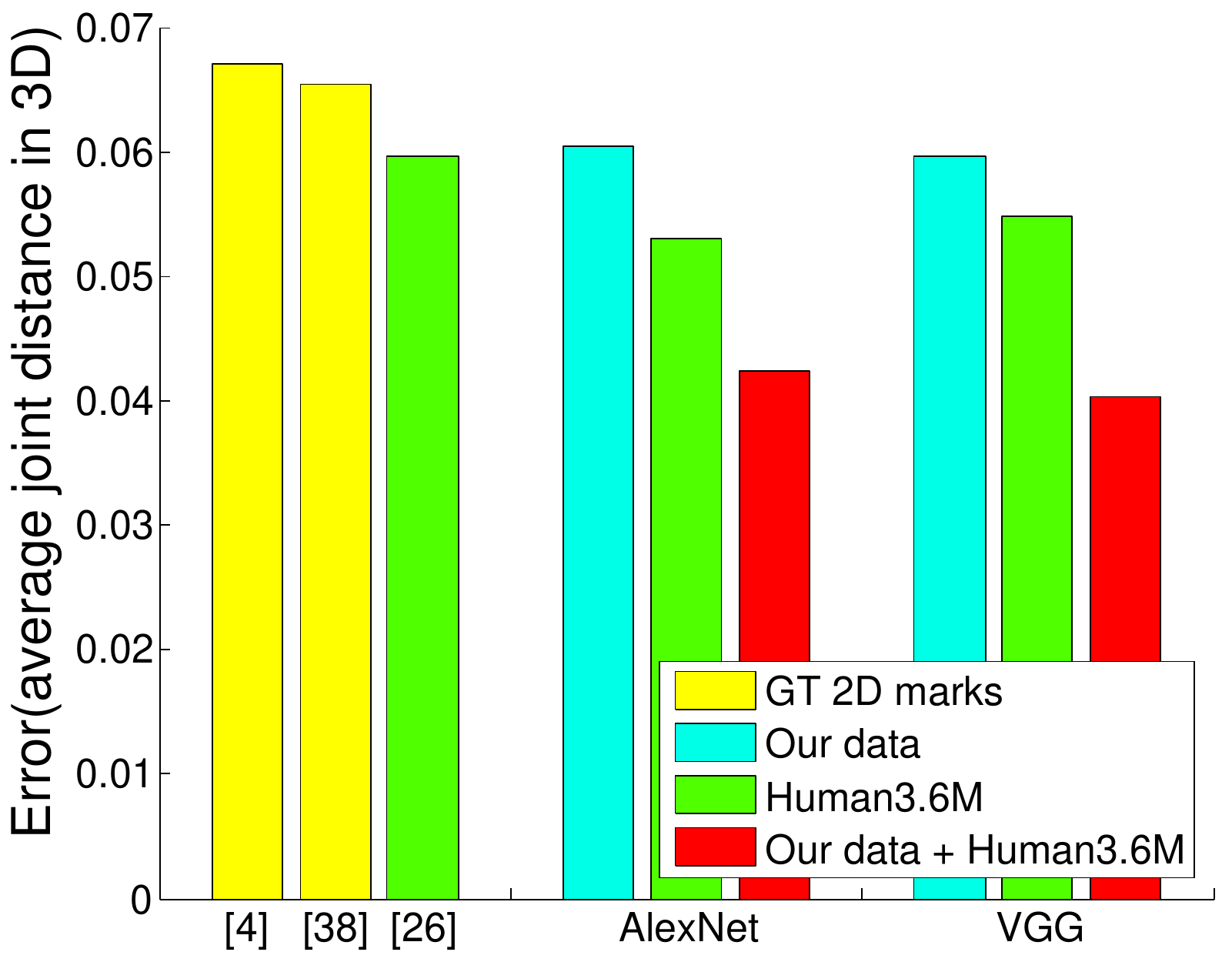}
		\includegraphics[width=0.4\linewidth]{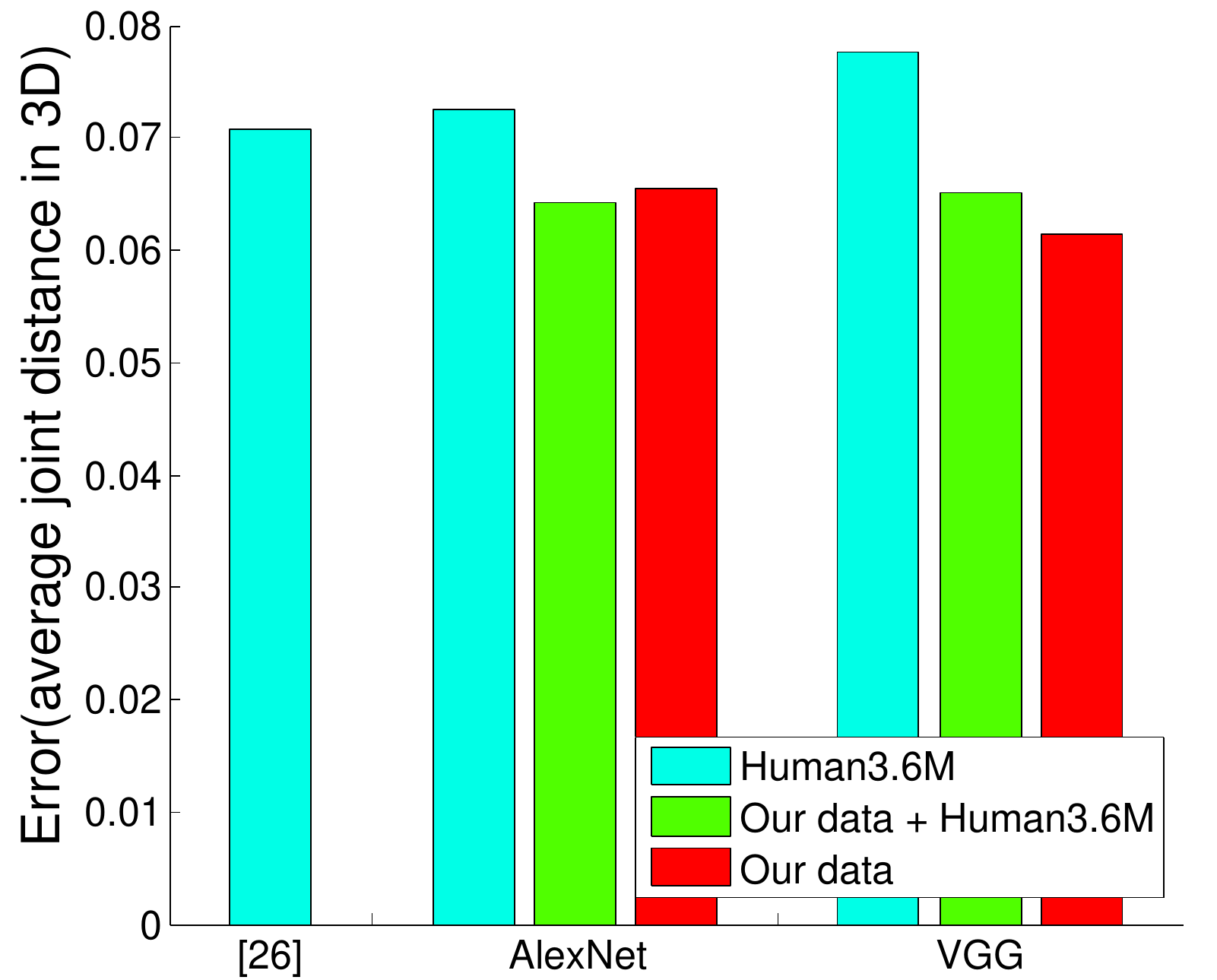}	
	\end{center}
	\vspace{-0.4cm}
	\caption{3D pose estimation evaluated on Human3.6M (left) and our Human3D+ (right). To better compare with different models and training data, we use bar graph to display the error. Various deep learning models (Li and Chan~\cite{li20143d}, AlexNet and VGG) trained on our data, Human3.6M, or a mixture of them are evaluated. We also compare against Ramakrishna et al. \cite{ramakrishna2012reconstructing} and Akhter et al. \cite{akhter2015pose} (left). To compare the generalizability of models trained on Human3.6M and our synthetic data, we evaluate these networks on a new dataset --- Human3D+ (right). }
	\vspace{-0.5cm}
	\label{fig:comparision_3d}
\end{figure}

\subsection{Evaluations on 3D Pose Estimation Task}
\label{sec:3d_pose}


Since the focus of this paper is on the generation of the training data, and it is not our intention to advocate a new network architecture for pose estimation, we test the effectiveness of our data using ``off-the-shelf'' image classification CNNs. Specifically, we adapt both AlexNet and VGG for the task of human 3D pose estimation by modifying the last fully connected layer to output the 3D coordinates, appended with an Euclidean loss, and fine-tuning all the fully connected layers to adapt them to the new task.

More specifically, our synthesis process outputs a large set of images $ \{\mathcal{I}_i\}$, each associated with a vector $\mathcal{P}_i \in \mathbb{R}^{45}$: the 15 ground truth 3D joint positions (in camera coordinates). The vector $\mathcal{P}_i$  defines the relative spatial relationships between the 3D joints (the human 3D pose), and also the camera viewpoint direction relative to a canonical human coordinate system (e.g, from which side of the human the camera is looking at it). We normalize the joint coordinates such that the sum of skeleton lengths is equal to a constant. We train the CNNs to estimate the 3D pose from a single input image. That is, given an image with a full human subject visible in it, the CNN yields 3D joint positions in camera coordinates. We denote the joint predictions as ${\mathcal{P}}^{'}_i \in \mathbb{R}^{45}$. We measure the prediction error with an Euclidean loss: $E = \sum_i {{\parallel \mathcal{P}_i - {\mathcal{P}}^{'}_i \parallel}_2} $.

\begin{figure}[t]
	\vspace{-0.5cm}
	\begin{center}
		\includegraphics[width=0.4\linewidth]{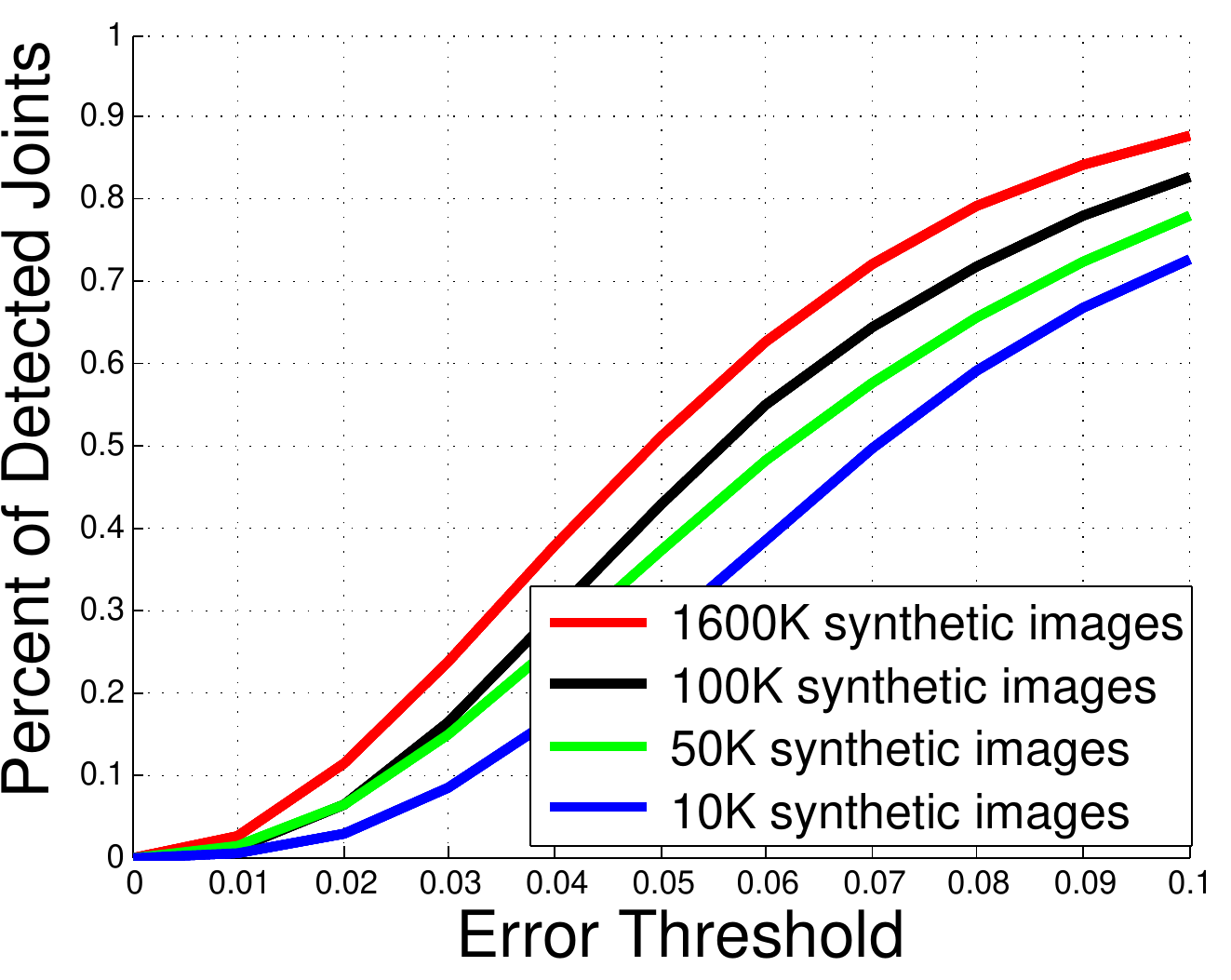}
		\includegraphics[width=0.4\linewidth]{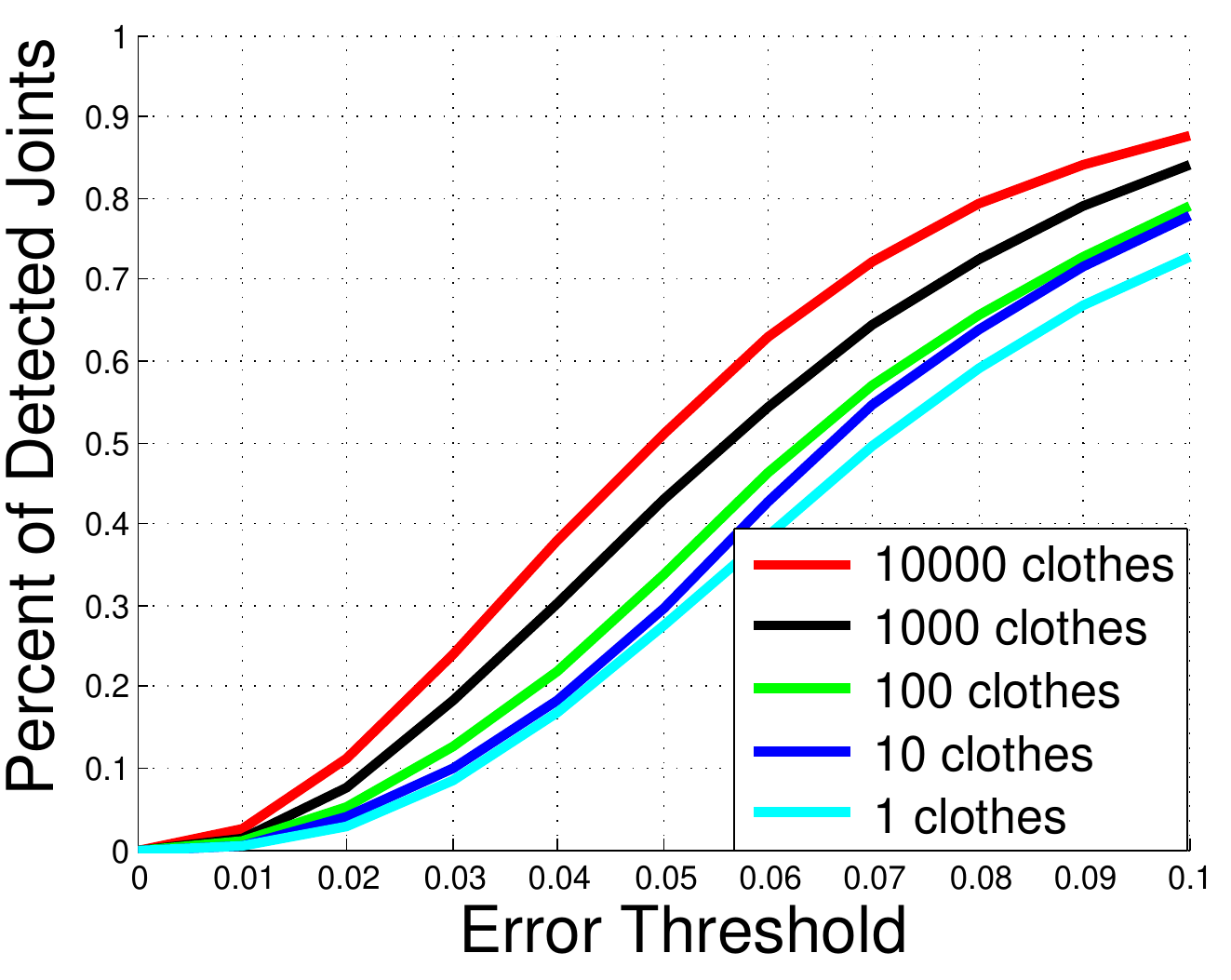}	
	\end{center}
	\vspace{-0.4cm}
	\caption{3D pose estimation performance increases with the size of the synthetic training set (left), and the number of different clothes textures used (right).}
	\vspace{-0.5cm}
	\label{fig:scalability}
\end{figure}

We compare the perofrmance of our simple adaptions of AlexNet and VGG with Li and Chan~\cite{li20143d}. Li and Chan train six models on different actions. We test all the testing images by their six models, and select the best one. Both our adaptions and Li and Chan output poses that are given in camera view. We first normalize and align the estimated 3D poses towards the ground truth, and then compare the results by plotting the percentage of detected joint points when different error thresholds are used against the ground truth annotations.

We train these three networks (VGG, AlexNet, and Li and Chan) on three training image datasets (our synthetic images, Human3.6M, and their mixture), and evaluate their performance on two evaluation datasets (Human3.6M and Human3D+). The performance of these variants is plotted in Figure~\ref{fig:comparision_3d}.

Several interesting observations can be made from the comparisons in Figure~\ref{fig:comparision_3d}. First, training on Human3.6M leads to over-fitting. While the models trained on Human3.6M apparently perform comparably or better than those trained on our synthetic images, when tested on Human3.6M (Figure~\ref{fig:comparision_3d} left), they perform less well when tested on Human3D+ (Figure~\ref{fig:comparision_3d} right), which is more varied than Human3.6M. Another evidence of the over-fitting is that VGG, which is generally considered to have larger learning capability than AlexNet, performs worse than AlexNet when trained on Human3.6M and tested on it (Figure~\ref{fig:comparision_3d} left), since it suffers from stronger over-fitting due to its larger learning capability. Second, it is clear that training with our synthetic data, rather than Human3.6M, leads to better performance on Human3D+ images, which exhibit richer variations (Figure~\ref{fig:comparision_3d} right). This shows a clear advantage of our synthetic images. Third, our synthetic images, when combined together with Human3.6M in the training, consistently improve the performance on both Human3.6M. This is an indication that our synthetic images and real images have complementary characteristics for the training of CNN models. We suspect our synthetic images cover larger pose space and texture variations, while Human3.6M images still have some characteristics, e.g. the realism, that are closer to real images than our synthetic images.

\begin{figure}[t]
	\vspace{-0.5cm}
	\begin{center}
		\includegraphics[width=0.4\linewidth]{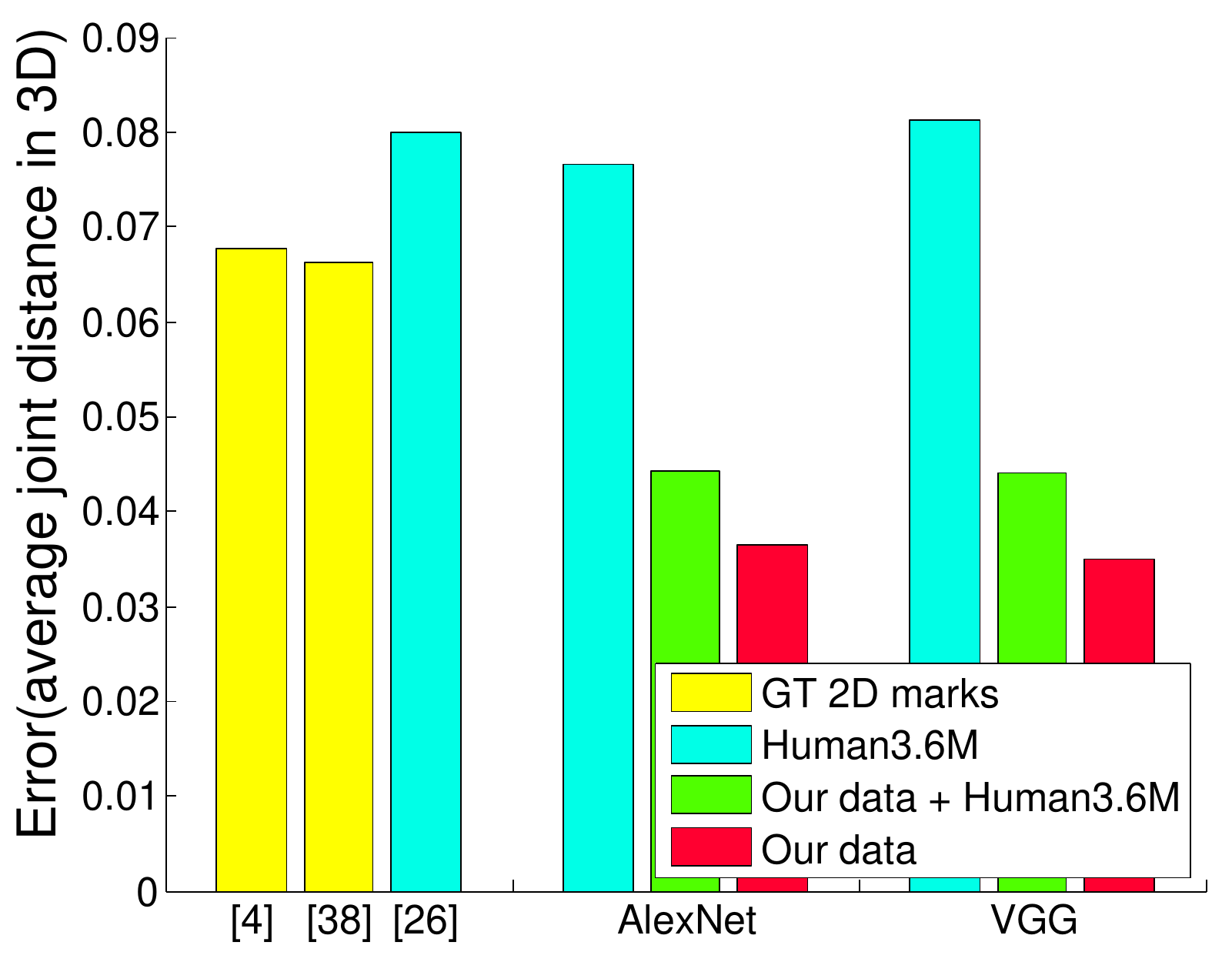}			
	\end{center}
	\vspace{-0.4cm}
	\caption{Performance of various models evaluated on our synthetic data.}
	\vspace{-0.5cm}
	\label{fig:on_synthetic}
\end{figure}

To get a better reference of the performance, we also compare against the methods~of Ramakrishna et al. \cite{ramakrishna2012reconstructing} and Akhter et al. \cite{akhter2015pose} which reconstruct a 3D pose from 2D joint locations. We found these methods to perform significantly worse, even when provided with the ground truth 2D poses (Figure~\ref{fig:comparision_3d} left)\footnote{Due to the technical limitation of~\cite{mocap_noitom}, the ground truth 2D poses are not available in Human3D+, thus this experiment could not be done on Human3D+.}. This is not suprising, as these methods take only the 2D joint positions as input, while ignoring the appearance. In contrast, CNN models effectively consume all the information in the input images.

\subsection{Training with Domain Adaptation}
\label{sec:domain_adaptation_results}

\begin{figure}[t]
	\vspace{0.2cm}
	\begin{center}
		\includegraphics[width=0.42\linewidth]{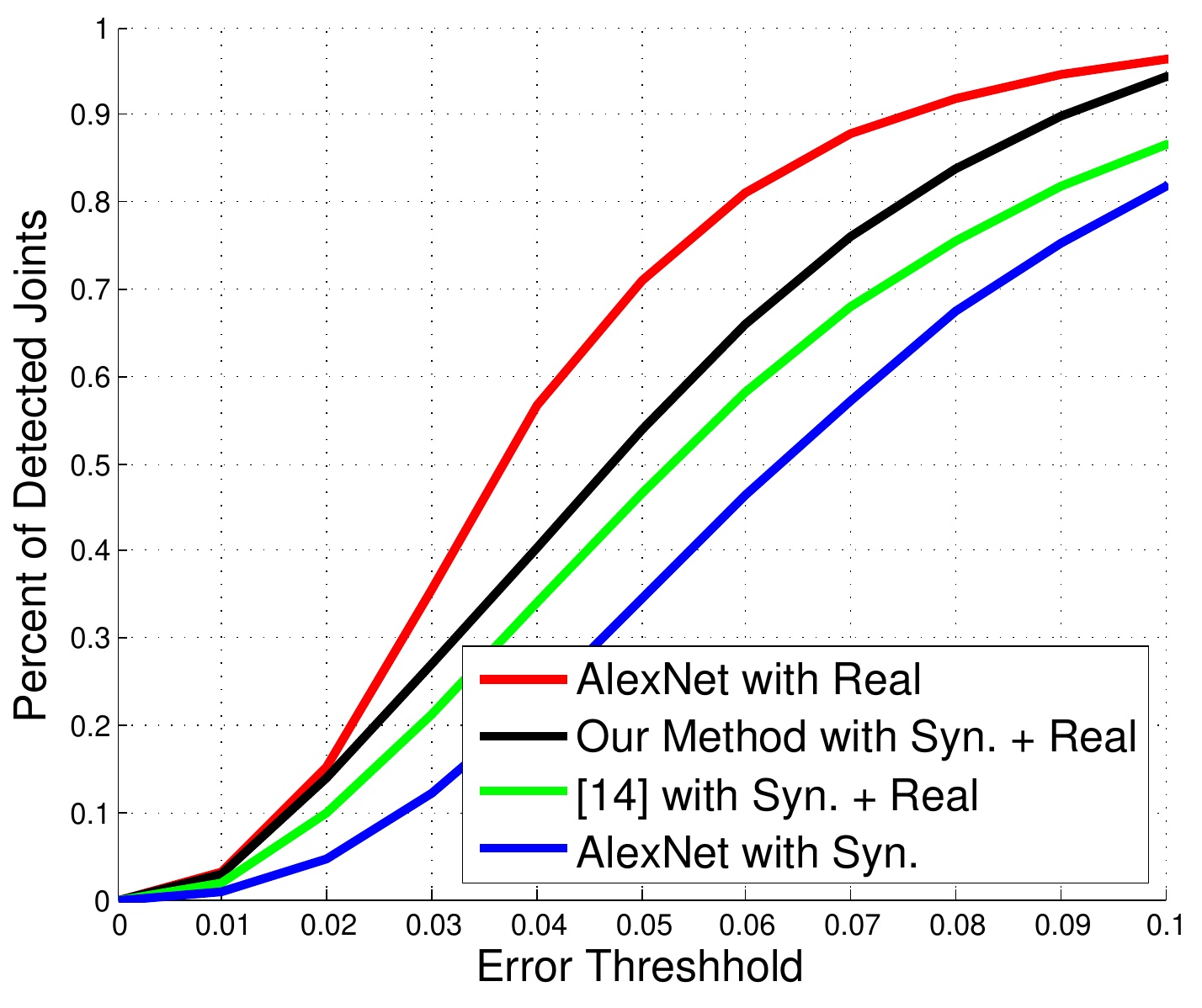}
		\includegraphics[width=0.4\linewidth]{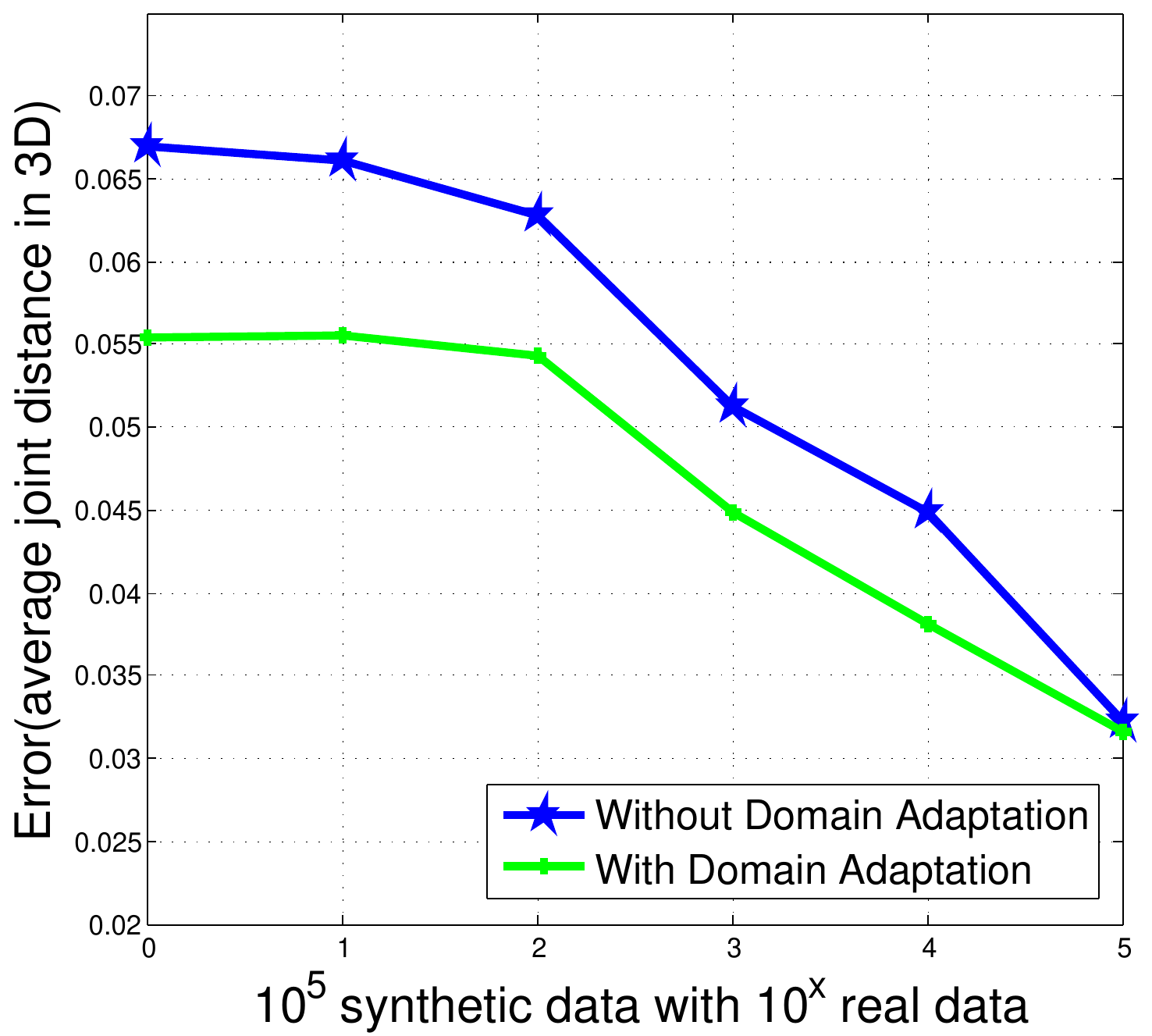}	
	\end{center}
	\vspace{-0.4cm}
	\caption{We show that our domain adaptation method can significantly improve 3D pose estimation performance (left), and the improvement is prominent especially when minimal amount of real images are available (right).}
	\vspace{-0.5cm}
	\label{fig:domain_re}
\end{figure}

We have shown that our synthetic images can benefit CNN model training on vanilla networks, when used alone, or together with real images with annotations in Section~\ref{sec:3d_pose}. In this section, we show that our synthetic images can benefit the domain adaptation network proposed in Section~\ref{sec:domain_adaptation}.

Since the annotations of real images are not required for the training of domain adaptation network, it is extremely useful for making best of synthetic images. We show that synthetic images trained with domain adaptation network performs significantly better than that on vanilla networks (see Figure~\ref{fig:domain_re} left). Also, we show our domain adaptation networ is better than ~\cite{ganin2015unsupervised}, since our method can better mixup the features extracted from synthetic and real images.

To show the effectiveness of synthetic images, we train our domain adaptation network with 100,000 synthetic images and 10, 100, 1000, 10,000 or 100,000 real images \emph{with} pose annotations, plus 50000 real images \emph{without} pose annotations as domain guidance, and compare with AlexNet trained with the same amount of images, without any guiding real images. As shown in Figure~\ref{fig:domain_re} (right), it is clear that our domain adaptation network can better utilize our synthetic images. Unsurprisingly, the benefit brought by the domain adaptation is prominent especially when minimal amount of real images are available.

\subsection{Parameter Analysis}
\label{sec:control_experiments}

\paragraph{Importance of scalability.} To investigate how important the number of synthetic images is for the 3D pose estimation performance, we train the same models using different synthetic training set sizes and report their performance in Figure~\ref{fig:scalability} (left). It is clear that increasing the number of synthetic images improves the performance of CNN models.

\begin{figure}[t]
	\vspace{-0.5cm}
	\begin{center}
		\includegraphics[width=0.8\linewidth]{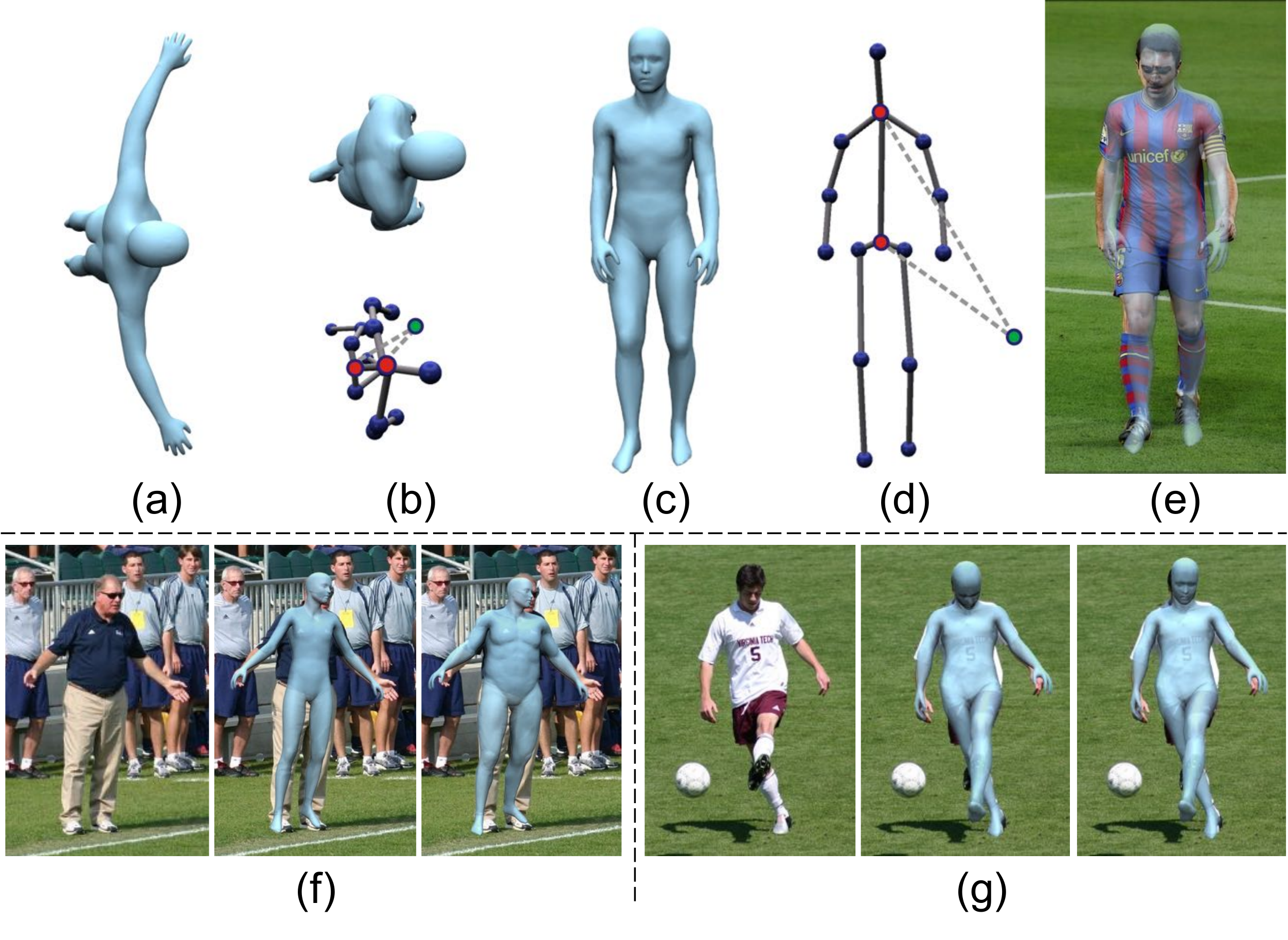}
	\end{center}	
	\vspace{-0.5cm}
	\caption{3D human reconstruction from single image. The SCAPE model in rest pose (a), can be articulated to (b) according to the pose estimated from the image (d). The rigid transformation between (b) and (d) can be computed from corresponding joints to align (b) to the human in the image (c). The reconstruction is visualized in (e). 
		}
	\vspace{-0.5cm}
	\label{fig:model_fitting}
\end{figure}

\paragraph{Importance of texture variability.} Similarly, we also study the impact of the number of different clothes textures used for ``dressing'' the human 3D models in Figure~\ref{fig:scalability} (right). Note that the richness of the clothes textures also plays important role in the overall performance, thus it is critical for the texturing steps to be as automatic as possible. In our case, only a modest amount of user input is required in the clothes images collection step, which actually can be further automated by collecting images from online clothing shops, or by a classifier trained for this task.

\paragraph{Evaluation on synthetic images.} To better understand the influence of our data for networks' generalizability, we test the various deep learning models on our synthetic data. The results are summarized in Figure~\ref{fig:on_synthetic}.
We see that the performance gap between models trained on real images (cyan) and on our synthetic images (green) is much more notable than in Figure~\ref{fig:comparision_3d}. It implies that, when the test data and the training data are from different sources, models trained on Human3.6M perform worse than those trained on our synthetic data. This asymmetry in the gaps is also another indication that our synthetic images have more variations than that in Human3.6M --- data with less variation is more likely to result in a model that performs well on itself but bad on new data.


\subsection{3D Reconstruction}

\begin{figure}[h]
	\vspace{-0.5cm}
	\begin{center}
		\includegraphics[width=0.8\linewidth]{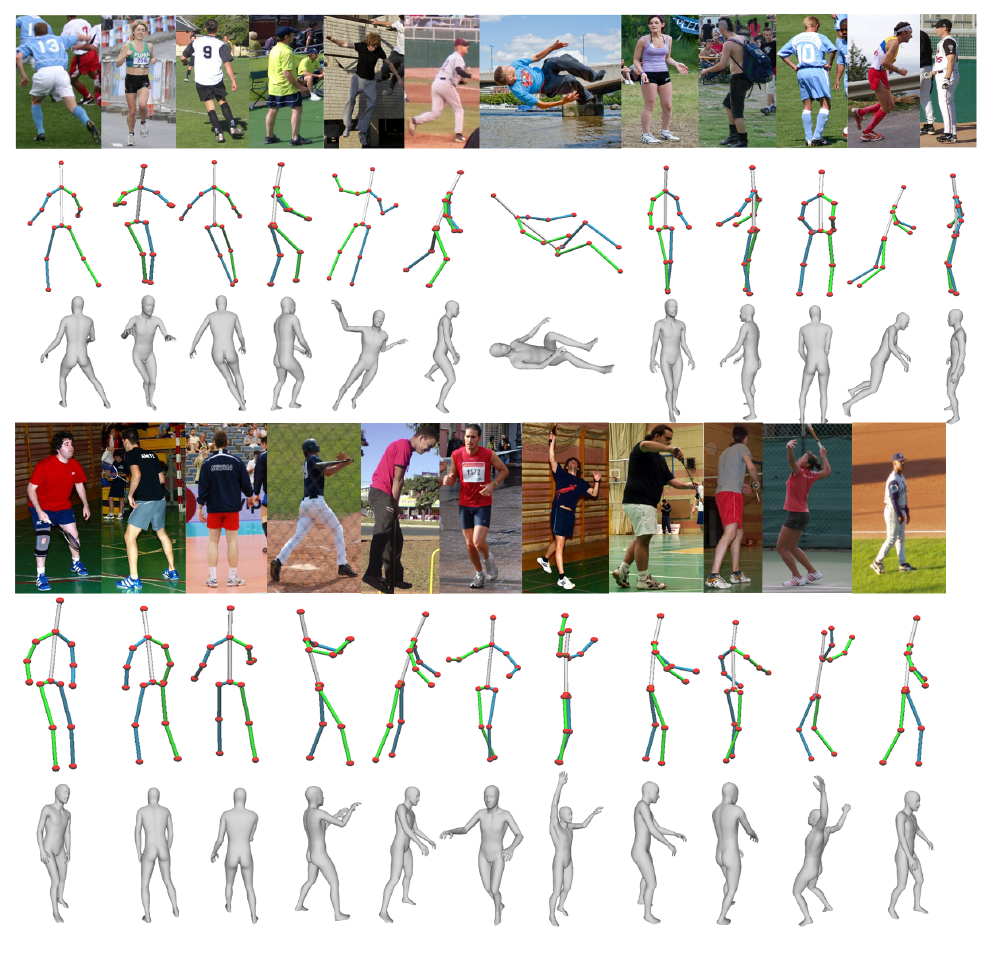}
	\end{center}s
	\vspace{-0.5cm}
	\caption{A sample of 3D human reconstruction from single image results, based on a 3D pose estimation model train on our synthetic images.}
	\vspace{-0.5cm}
	\label{fig:results}
\end{figure}

Human 3D pose estimation from a single image is an important step towards human 3D reconstruction from a single image. The estimated 3D pose can be used to articulate a SCAPE model, as well as align it to the human in the image. The articulated aligned model can already serve as a feasible 3D reconstruction, as shown in Figure~\ref{fig:model_fitting}, and more in Figure~\ref{fig:results}. However, more faithful 3D reconstruction requires recovering additional 3D properties from the input image. As shown in Figure~\ref{fig:model_fitting} (f) and (g), body shape and gaze also play important roles. Similarly to pose, such 3D properties can be hard to annotate, but come free from the synthesis pipeline. We believe our work will encourage more research along these directions.

\section {Future Work and Conclusions}
\label{sec:future_work_and_conclusion}
\label{sec:conclusions}

Training data for inferring 3D human pose is costly to collect. In our system to synthesize training images from 3D models, the association between the images and the 3D ground truth data is available for free.  We found the richness of the clothing textures and the distribution of the poses to be of particular importance. However, constructing a model for realistic clothing synthesis from scratch is a difficult challenge in itself, so we propose instead to sidestep this challenge by transferring clothing textures from real images. We show synthetic images can be better utilized with domain adaptation. We show that the CNNs trained on our synthetic data advance the state-of-the-art performance in the 3D human pose estimation task.
We plan to make all of our data and software publicly available to encourage and stimulate further research.

\section {Acknowledgements}
We thank the anonymous reviewers for their valuable comments. This work was supported by the National Key Research \& Development Plan of China, No. 2016YFB1001404.

{\small
\bibliographystyle{ieee}
\bibliography{egpaper_final}
}

\end{document}